\documentclass[a4paper]{article}

\usepackage{lineno}
\modulolinenumbers[5]

\usepackage{graphicx}
\usepackage{amsmath,amsfonts,mathrsfs,amssymb}
\usepackage{bm}
\usepackage{algorithm}
\usepackage[noend]{algorithmic}
\usepackage{hyperref}
\usepackage{subfig}
\usepackage{color}
\usepackage{rotating}
\usepackage{multirow}
\usepackage{placeins}
\usepackage{soul}
\usepackage{marvosym}

\usepackage{graphicx}
\usepackage{amsmath}
\usepackage{longtable}
\usepackage{color}

\usepackage{makecell}
\usepackage{lscape}
\usepackage{mathtools}
\usepackage{caption}
\usepackage{subfig}
\usepackage{float}

\usepackage{color}
\usepackage{longtable}

\usepackage{color}
\usepackage{longtable}
\usepackage{caption}
\usepackage{makecell}
\usepackage{lscape}
\usepackage[margin=1in]{geometry}

\newcommand{\sep}{ }

\begin{document}

\author{Jasmin Bogatinovski\textsuperscript{1,2,3\Letter}, Ljup\v{c}o Todorovski\textsuperscript{1,4}, Sa\v{s}o D\v{z}eroski\textsuperscript{1,2}, Dragi Kocev\textsuperscript{1,2,5\Letter} \\
\textit{\textsuperscript{1}Jo\v{z}ef Stefan Institute, Ljubljana, Slovenia} \\
\textit{\textsuperscript{2}Jo\v{z}ef Stefan IPSchool, Ljubljana, Slovenia}\\
\textit{\textsuperscript{3}Dept. of Distributed Operating Systems, TU Berlin, Germany}\\
\textit{\textsuperscript{4}Faculty of Mathematics and Physics, University of Ljubljana, Slovenia}\\
\textit{\textsuperscript{5}Bias Variance Labs, Ljubljana, Slovenia}\\
\{name.surname\}@\{tu-berlin.de\} \{ijs.si\}
}
\date{\today}
\title{Explaining the Performance of Multi-label Classification Methods with Data Set Properties}

\maketitle

\begin{abstract}
Meta learning generalizes the empirical experience with different learning tasks and holds promise for providing important empirical insight into the behaviour of machine learning algorithms. In this paper, we present a comprehensive meta-learning study of data sets and methods for multi-label classification (MLC). MLC is a practically relevant machine learning task where each example is labelled with multiple labels simultaneously. Here, we analyze 40 MLC data sets by using 50 meta features describing different properties of the data. The main findings of this study are as follows. First, the most prominent meta features that describe the space of MLC data sets are the ones assessing different aspects of the label space. Second, the meta models show that the most important meta features describe the label space, and, the meta features describing the relationships among the labels tend to occur a bit more often than the meta features describing the distributions between and within the individual labels. Third, the optimization of the hyperparameters can improve the predictive performance, however, quite often the extent of the improvements does not always justify the resource utilization.
\end{abstract}
Machine learning \sep Meta learning \sep  Multi-label classification \sep Meta knowledge \sep Hyperparameter tuning

\section{Introduction}\label{sec:introduction}

A successful application of a learning method on a practical problem is preconditioned on knowledge about both the learning method and the domain of the problem~\cite{Alex, yeast, KocevNature}. 
The learning methods are often conceptualized into different families of methods.
Each family includes methods that make a certain set of assumptions about the learning problem (better yet the training data). It is often beyond trivial to find which set of assumptions represent the best fit for a given problem, making the selection of the appropriate method (or family of methods) a complex decision. Following the Law of Conservation of Generalization~\cite{Wolpret1996} it is not even clear which method or at least which family of methods is preferred for the given practical problem. To overcome the challenge in successful application of machine learning on practical problems, a plethora of areas in machine learning have emerged -- AutoML~\cite{Hutter}, meta-learning~\cite{STATLOG, METAL, Vilalta2002,Brazdil2009,Lemke2015:jrnl,Brazdil2018:jrnl,Soares1,Joauqin2018,oldyGoldy}, transfer learning (inductive transfer)\cite{transferLearning}, learning to learn~\cite{Thurn}, life-long learning, continual learning~\cite{Ring97}, multi-task learning~\cite{Caruana1997}, domain adaptation~\cite{domainAdaptation} and domain generalization~\cite{domainGeneralization} -- with sustainable overlaps among them. 
Common to these areas is that they utilize a set of problems emerging from various domains (referred to as base-learning problems) or different realization of the same problem, (referred to as varieties/tasks). Furthermore, the approaches from the aforenamed areas aim towards creating a joint abstract representation (meta knowledge) of the varieties, which can be utilized in a life-long learning machine as defined by Thurn~\cite{Thurn, Thurn1998isNthingEasierToLearn}, or to perform the hard task of algorithm selection~\cite{Rice}, among other applications.

The meta knowledge can be observed as empirical manifestation of the varieties and a footprint of the behaviour of the learning methods for the learning task. One of the goals of meta learning as an area in machine learning is to determine the exact properties of the learning task that make a method suitable for a specific meta region (in a space described by meta features), while expressing the components contained by the method that contributes to it dominating specific regions in the metaspace~\cite{Vilalta2002}.
Despite its main intent as a strong prior for improving the generalization performance of the learning methods, the meta knowledge that reflect comprehensibility~\cite{GirardCarier1997} can also be used in gaining valuable insights about the properties of the learning method or learning task itself~\cite{Aha1992,Brazdil1995,Domingos2000,Mantovani2015}. 

Examples of comprehensible meta knowledge gained through meta learning include: Domingos~\cite{Domingos2000} argues that empirical evidence across many problems suggests that for the zero-one loss~\cite{Duda2001}, C4.5~\cite{Quinlan1993} does not benefit from tuning the hyper-parameter confidence level. Tibshirani et. al.~\cite{ElementsOfStatisticalLearning} argue that $k$-NN for regression performs well when the number of features of the data does not exceed four and there are more than 1000 samples, with sufficient support, so the region is assumed to be dense enough. Recently, Ridd and Giraud-Carrier~\cite{Girard} show that the performance of a 1NN classifier is sufficient to predict when hyper-parameter optimization is likely to improve the performance of a classification method on a given problem. Since 1NN is a measure for the density of the instances in the problem space, the high values show that when the input region is dense, one may not benefit from tuning the hyper-parameters. Similarly, Mantovani et al.~\cite{Mantovani2015} show that if the datasets are very imbalanced tuning of the parameters of SVM is most likely not beneficial. 

In this work, under the term comprehensible meta knowledge we understand any statistics (as a function of training data) calculated on the problem that reflects the properties of the learning task and can be assigned meaning by a human. By inducing a relation between the properties of the problems and the performance of ML methods, we aim to deduce comprehensible meta knowledge.
This form of meta knowledge is of sustainable value for both the practitioners and experts for several reasons. First, it provides a complementary view over the theoretical studies of the methods, promoting or constraining their practical applicability. Second, it points out the potential limitations of the methods, ideally providing guidelines for improvement~\cite{Domingos2000}. Third, it allows studying the behaviour of methods in terms of the main properties of the learning task. Therefore, it translates to an empirical study of the practical challenges of the learning task.  Finally, it empowers the practitioners with better access and understanding of the behaviour of the methods under the practically faced limitations, therefore, making an informed decision about the method selection. Ultimately, this can lead to better acceptance and increased trust in the adoption of machine learning methods by the practitioners, reducing their time and efforts for better understanding the methods thereof focusing more on the problem at hand. 


While the studies related to the deduction of meta knowledge from comprehensible studies are dominating the learning tasks of a single target, the works on other learning tasks are rare or non-existent. The reason for the reformulation of a problem under different learning task often resides in improved predictive and prescriptive performance~\cite{Jurica, KocevNature, Leskovec, Blazh}. One such example is the task of multi-label classification. 

Multi-label classification (MLC) is a machine learning task that falls into the structured output prediction paradigm. The main property of MLC is that each sample is annotated with multiple labels simultaneously. The goal is to learn model(s) discerning the relevant from the irrelevant labels for a given data point. It is a reformulation of the binary classification task where instead of one target, one aims to predict several targets at once. MLC introduces novel properties and challenges that do not emerge in the single-target scenario, for example, high dimensional target space, extreme imbalance of the target space and the dependence between the targets. 

Addressing problems as a MLC task provides improved performance on many practical problems~\cite{Madjarov2012, Herrera2016}. Therefore, it is an attractive area of research, evident by the constant increase of the number of methods and problems~\cite{Bogatinovski2019} (e.g., for the year 2019 the SCOPUS database (\url{https://www.scopus.com/}) registered around 1000 papers on the topic of MLC). Furthermore, the MLC task can be used to solve a variety of practically relevant tasks from different domains~\cite{Madjarov2012,Herrera2016}, including life sciences and medicine (e.g., gene and protein function prediction, drug modes of action, drug repurposing, patient status prediction), environmental sciences (e.g., habitat models, compounds' toxicity), text classification (e.g., tag suggestion, news articles, web pages, patents, e-mails, bookmarks, recipes), semantic annotation of images and videos (e.g., image annotation, news clips, movies, species identification from recordings, music evoking emotions), directed marketing and others.

Following the accessible meta knowledge for the single target classification and regression tasks, in this work, we connect the dataset properties with the performance evaluation of the MLC methods to deduce a comprehensible meta knowledge for the MLC task. More specifically, we investigate how the properties of the MLC task constrain the behaviour of the learning methods for the problems, we describe with meta features. For this purpose, we exploit the results from the largest study of MLC methods to date, analyzing the performance of 26 MLC methods on 40 MLC datasets using 18 predictive performance and two efficiency measures. 
The used data sets vary significantly in terms of the number of examples, features and labels. Furthermore, the used methods cover a large spectrum of the most widely established MLC methods from the literature. By considering such a variety of data sets and methods, we perform an in-depth meta-learning study and analysis of the MLC task.

The main contributions of this study can be summarized as follows:
\begin{enumerate}
    \item  We extensively overview the meta features for MLC and propose a taxonomy of meta features, which facilitates the pursuit of comprehensible meta knowledge for the MLC task.
    \item By describing the landscape of MLC data sets with a clustering tree, we identify the potential of the meta features for MLC to identify groups of MLC datasets sharing similar meta feature description and similar method performance.
    \item By relating the meta features with the predictive performance of individual and groups/families of MLC methods, we discriminate the domain of expertise of the best performing MLC methods and their expected performance.
    \item We study the influence of the tuning of the hyperparameters of the methods on the predictive performance of the models, identifying scenarios where the need for performing hyperparameter tuning is favourable against reliable baselines.
\end{enumerate}

The remaining of this paper is organized as follows. In Section 2, we give the relevant background from both MLC and meta learning. In Section 3, we propose a taxonomy of meta features. We also describe the experimental study as well as the meta-learning scenarios considered here. Section 4 discusses the potential of the meta features to describe the space of MLC datasets. Section 5 relates the meta features with the predictive performance of the MLC methods. Section 6 answers the question when the hyperprameter tuning is improving the performance over reliable defaults. In Section 7 we give an overview of the main findings enriching the MLC knowledge corpus. The conclusions and further directions are provided in Section 8.

\section{Background}
In this section, we give the background information for the study. We first provide a formal definition of the MLC learning task. Second, we describe the families of methods and the evaluation measures as necessities for successful discussion. Finally, we present the related work on meta learning for MLC. Due to space limitations, all the additional details for the used datasets, the definitions of the evaluation measures and the descriptions of the used methods, together with the appropriate abbreviations, can be found in the Supplementary material.

\subsection{Multi-label Classification}
MLC is a machine learning task where the goal is to predict the subset (out of a predefined set) of labels that are relevant for a given example~\cite{Madjarov2012,ZhangTPAMI,Herrera2016,Moyano2018}. By doing so, for each example a bipartition of the relevant and the non relevant labels is defined. Formally, the MLC task is defined as follows~\cite{Madjarov2012}: Given a $d$-dimensional input space $\mathcal{X}$, an output space of $|L|$ labels $\mathcal{Y}=\{l_{1}, l_{2} \dots l_{|L|}\}$, and a training set of $N$ samples defined as $ D=\{ (\mathbf{x_{1}}, Y_{1}) \dots (\mathbf{x_{i}}, Y_{i}) \dots (\mathbf{x_{N}}, Y_{N})\}$, where  $\mathbf{x_i}=(x_{i1}, x_{i2} \dots x_{id}), \forall{\mathbf{x_i}}\in\mathcal{X}$ is the input description of the sample $i$, $Y \subseteq \mathcal{Y}$ is called labelset, $(\mathbf{x_i}, Y_i)$ is multi label sample, and $|L|>1$, the goal of MLC is to learn a function $h$, such that $h: \mathcal{X}_{|D} \mapsto{2^\mathcal{Y}}$ while optimizing a given quality criteria $q$.

\subsubsection{Taxonomy of Methods}
The methods for addressing the task of MLC are typically divided into two groups: \emph{problem transformation} (PT) and \emph{algorithm adaptation} (AA) methods~\cite{Gibaja2014}. PT methods transform the task into multiple simpler (typically single target) tasks and apply well-established single target methods for learning predictive models on them. Furthermore, we distinguish methods transforming the task into several binary classification tasks (i.e., Binary Relevance - $PT.BR$) or single multi-class classification task (i.e., Label Powerset - $PT.LP$). 
AA methods adapt single target methods to handle MLC directly. The adaptation includes modification of the methods by changing the underlying assumptions made by the single target methods to fit the multiple targets simultaneously. 

\subsubsection{Taxonomy of Evaluation Measures}
While the evaluation of the single target methods is straightforward, the MLC methods have an additional dimension one should consider; the combination of the performance across the multiple labels. The assessment of the performance of an MLC method relies on the type of predictions it produces: relevance scores per label or a bipartition. 
If the method produces relevance scores per label, two performance scores groups exist; threshold independent and ranking-based measures. For methods outputting relevance scores, one can apply special postprocessing techniques to produce bipartitions~\cite{Reem2014}.

In the case of bipartitions, the evaluation measures are grouped into example-based and label-based measures. The latter further is split into micro and macro-averaged performance scores, based on how the joint statistics for all labels are calculated. If the statistic is calculated jointly from the contingency matrix we refer to it as micro, while if the scores calculated for individual label and then averaged into a single value we refer to it as a macro aggregation of the performance scores. 

The different methods might be more biased towards optimizing a given evaluation measure than other methods. For example, the methods predicting label sets have favourable evaluation using example-based measures, while the methods predicting each label with a different model and combine the predictions have favourable evaluation using macro-averaged measures. Hence, for an unbiased view on the performance of the MLC methods, one needs to consider multiple evaluation measures. 





\subsection{Related Work}
Despite the important practical relevance of the MLC task, the available work on explaining the predictive performance of MLC methods with dataset properties, particularly the aspect of investigating the comprehensible meta knowledge for MLC, is very scarce. To begin with, an overview of available meta features for MLC is given in~\cite{Moyano2017}. The meta features are divided into six groups without providing semantic links among the groups. Therefore, the reasoning over the different meta features is more challenging. In Moyano et al.~\cite{Moyano2018}, the authors consider a meta-study of ensembles of MLC methods. They utilize four meta attributes selected by domain experts that reflect the dataset size as well as imbalance and dependency between the labels. The considered grouping of the datasets is univariate and with preselected thresholds over the features. The presented analysis is limited in scope and context concerning the number of considered datasets (20 datasets), the number of methods (12 methods) as well as the number of meta features. As such they do not provide a detailed landscape for the MLC task.

In another study, Chekina et al.~\cite{Chekina2011} investigate the question of the most suitable method for a new unseen MLC data set. Their study includes 12 MLC datasets. To elevate the problem of learning from a small number of examples, they use the original dataset in a data augmentation procedure. More specifically, the availble datasets are used as seeds for artificially generating additional data sets by the techniques of projecting, selecting and distorting the original data sets~\cite{Rokach2006}. This procedure results in a total of 640 datasets. Their comparative study includes seven single models and four ensemble methods. However, the meta analysis is conducted using two evaluation measures for the single models and two evaluation measures for the ensemble models. As meta learner adopted a k-nearest neighbour method (with k-2). The meta learner was able to identify the best performing method in 70-80\% of the cases (it varies depending on the selected evaluation measure). Moreover, they identified the most influential meta features contributing to certain decisions.  The conclusions from the study are limited from several aspects: (1) the statistical analysis of the performance is performed over examples that are not independent (recall that the 640 data sets are in essence 12 data sets that were manipulated), (2) the number of methods and data sets is small, and (3) the methods use the default values of the parameters without parameter tuning (the base classifiers are decision trees) thus limiting the expressiveness power of some methods concerning others.



The work presented here goes beyond the previous studies by a sustainable margin. First of all, the size of our study is an order of magnitude bigger than the other meta-learning studies (and the other experimental studies for that matter). Second, we perform parameter selection of the base predictive models (typically, support vector machines - SVMs)~\cite{Madjarov2012}. Third, we perform the analysis using multi-target prediction models: These can provide a unifying overview of the meta knowledge since a single model can describe the performance of a method across multiple evaluation measures. All in all, we address the task of meta learning for MLC with a comprehensive and rigorous empirical study.

\section{Meta Analysis Framework for MLC}

In this section, we first present the considered meta learning tasks related to the research questions of interest. Next, we describe the MLC specific meta features used in this study. Furthermore, we discuss the need for having interpretable predictive models to perform the meta-study. We then present the study that produced the experimental results. Finally, we give details about the specific implementation of the methods and experimental framework.

\subsection{Meta analysis Learning Scenarios}

In this study, we investigate three meta analysis scenarios\footnote{Note that the questions are formulated in the context of MLC, however, they can be applied to any other learning task.} to extract comprehensible meta knowledge for the MLC task: 
\begin{enumerate}
    \item {What is the potential of the meta features to describe the space of MLC datasets?}
    \item {Whether and how the meta features are related to the predictive performance of the MLC methods?}
    \item {Does tuning of MLC methods improves their predictive performance?}
\end{enumerate}

The first question aims to answer whether the used meta features can provide a concise description of the meta-space while completely ignoring any form of knowledge about the predictive performance of the methods. For this scenario, we focus on the descriptive power of the meta features and use them for clustering of the data sets. We later assign the predictive performance of the methods to each of the clusters, grouping the methods by their corresponding family. The analysis of the clusters reveals whether the family of methods have stronger performance for a specific cluster described by the meta properties. We conjecture that the top-performing MLC methods in each cluster show a preference for datasets with specific properties, thus leading to meta knowledge on the domain of expertise for each paradigm. Ultimately, this learning scenario measures what features are informative and should be applied to guide the practitioners when selecting a suitable method. Additionally, the experts can use this knowledge to investigate specific groups of datasets and reason how those properties can be utilized to improve the methods.


The second question directly relates the meta features with each of the predictive performance of the methods. It shows which meta features are of interest for which performance measure. Pairing the values of the relevant meta features on a new problem with the induced meta models, examines the expected performance of the method, within a predefined margin of error. The interpretable meta features and models allow to reason why certain methods perform good for data sets with certain meta-properties and not for others. We conjecture that using the meta features one can learn accurate meta models for predicting the performance of MLC methods on a given data set. To this end, we use compact multi-target regression trees~\cite{Blockeel1998, Kocev:Journal:2013} that can be readily transformed to meta knowledge on the performance of MLC methods. 

The third question addresses the benefit of hyperparameter tuning for MLC. The tuning of the hyper-parameters of the methods can be a computationally expensive procedure. Studies in machine learning show that the tuning often can result in marginal improvement over the default/recommended values for the hyper-parameters of the methods~\cite{Girard}. To verify the benefit of tuning, but more importantly, to provide guidance if tuning is expected to provide greater improvements on a given dataset, we construct a learning problem tailored to answer whether and when the tuned method is expected to produce better results than reliable baseline models. 

The experts can benefit from this information, in a way that allows constraining the hierarchical, heterogeneous multi-type search space in AutoML solutions e.g.~\cite{Hutter, Wever2018}. It is especially beneficial for tasks such as MLC, where there are up to three layers in the hierarchy of the hyper-parameters that should be adjusted. Following this, we analyze the benefit of hyperparameter tuning for individual MLC methods and by comparing the performance of two classes of MLC methods: the class of methods sensitive to the settings of hyperparameters that require tuning and the class of methods with strong default settings of the hyperparameters that do not need tuning \textit{reliable-defaults}. This question provides meta knowledge identifying the cases where hyperparameter tuning leads to improved predictive performance.

\subsection{Meta Features for MLC}
\begin{figure*}[!t]
\includegraphics[width=0.95\textwidth]{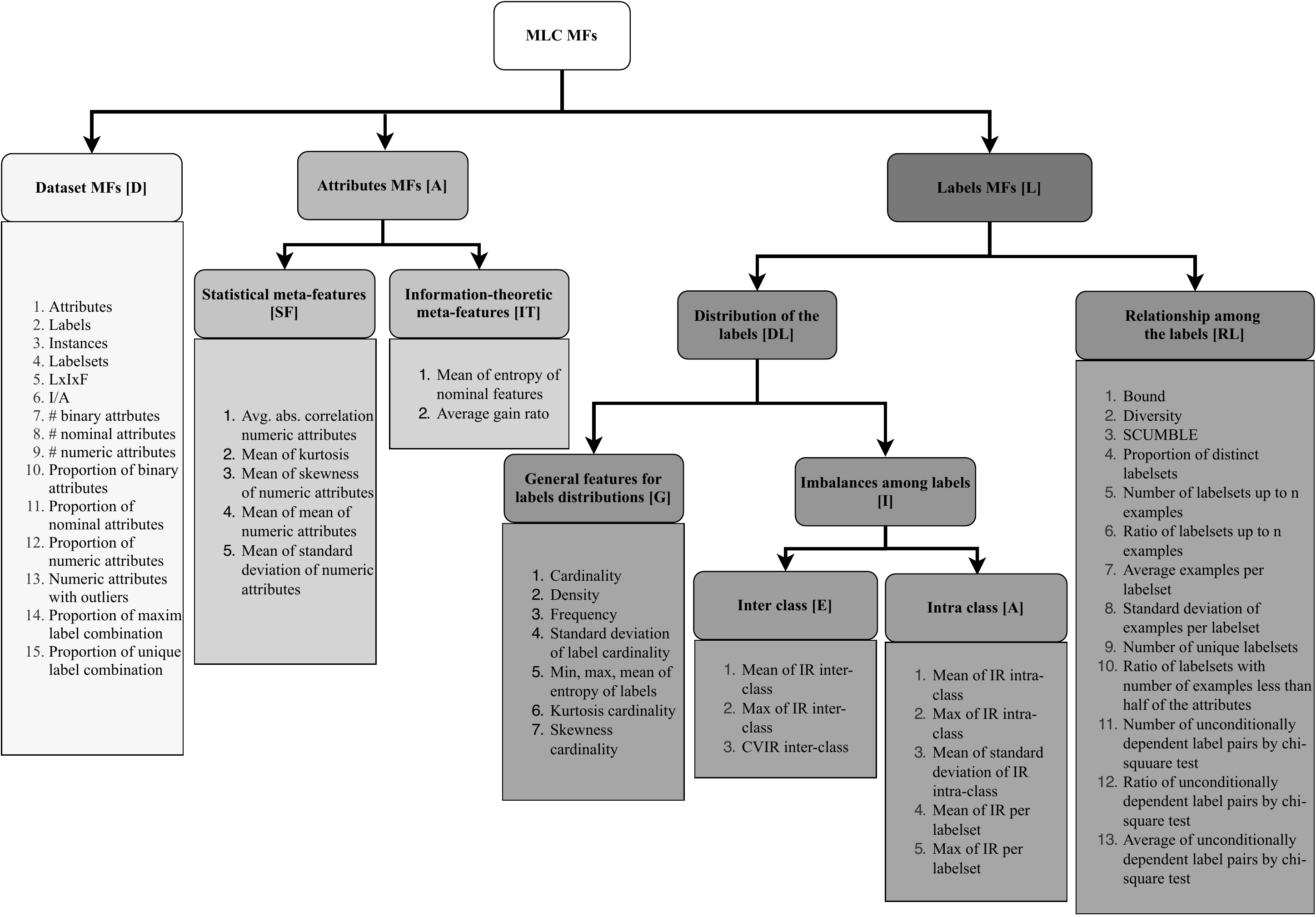}
\centering
\caption{A taxonomy of the meta features describing multi-label classification (MLC) data sets. For easier tracing, the meta features from the taxonomy are shortly labelled with hierarchical codes as follows: The first character is the group on the top level, the second character is on the second level etc. For example, \texttt{L.DL.I.A.4} represents the meta feature \texttt{Mean of IR per labelset} that belongs to the groups of features \emph{label meta features} (L),\emph{distribution of the labels} (DL), \emph{imbalancedness among labels} (I), \emph{intra class} (A), and is on the 4-th position within the last group of features.}
\label{sec3:taxonomy}       
\end{figure*}

There are many MLC-specific meta features that can characterize different aspects of the MLC data sets~\cite{Charte_2015, Moyano2017}. We analyze each of the available meta features and design a novel taxonomy to better organize and catalogue them (as illustrated in \figurename~\ref{sec3:taxonomy}). It allows us to facilitate the interpretation of the induced meta models. More specifically, it helps us to relate various features across the taxonomy to better and easier discover existing relationships among the meta features.

The left-most group of features describe the \emph{dimensionality} of the data sets, i.e., it includes meta features such as the number of features, instances, labels and their combination in terms of products and ratios.
The middle group of meta features describe \emph{statistical and information-theoretical} properties of the features/attributes of the data sets, such as the counts of categorical and numerical features, different ratios in terms of both the number of instances and number of the complementary data types, etc. 
The right-most group of meta features describes the label space by considering the \emph{distribution} of the labels and the \emph{relationship} among the labels.
The former group is subdivided into two groups i.e. general features representing the distribution of the labels and features representing the unbalancedness among the labels. This group of meta features includes the information on cardinality, density, frequency, kurtosis, skewness and entropy of the labels, as well as a variety of properties describing the inter and intra-label imbalance ratios.

The group of meta features describing the relationships among the labels consider different approaches to quantify the relationships and existence of different label sets in the data sets. It includes meta features such as the number of unique label sets, standard deviation among the examples per label sets, and meta features describing the pairwise dependency between the labels.


\subsection{Interpretable meta models}
In pursuit of comprehensible meta knowledge, the meta-model should be interpretable and able to express the amount of variance it decomposes while being robust to correlated meta feature dimensions. Owning to the small sample size typically encountered in meta learning studies, this consideration is crucial. Therefore, we use predictive clustering trees (PCTs)~\cite{Blockeel1998, Kocev:Journal:2013} for learning interpretable meta models. PCTs are a generalization of decision trees towards the task of structured output prediction (such as multi-target regression, multi-target classification, multi-label classification and their hierarchical variants). PCT is a flexible method allowing its usage in diverse learning scenarios: clustering, single target prediction and multi-target prediction. The latter helps to build simpler models that are interpretable, while still produces comparative predictive results with the single target models. Lundberg et al.~\cite{shap} show that tree models are more interpretable than linear models due to the model mismatch effect (i.e, linear models tend to be more influenced by interaction instead of the marginals of the interaction making them difficult for interpretation). Therefore, we find that tree models are more suited for the considered meta learning tasks. 

PCTs are learned using the standard top-down induction of decision trees algorithm. The utilized heuristic function minimizes the intra-cluster variance. PCTs, use specific instantiations of their heuristic and the prototype function to address the different tasks (clustering, single target or multi-target prediction). 

Namely, the specific meta-analyses related to the research questions can be executed by defining tasks and using the PCTs in an appropriate context. For example, using the meta features as both descriptive and clustering features one can use PCT to perform unsupervised divisive clustering. Relating the predictive power of several MLC methods with the meta features is a problem of multi-target regression. The descriptive variables are the meta features, while the targets are the performance scores. The single-target scenario is a special case of the multi-target scenario, where one considers just one target variable. Finally, by changing the heuristic function to information gain, one can use PCT method for performing classification (e.g. selecting the best performing method). 

\subsection{Experimental Study}

In this study, we perform meta learning over the experimental results from~\cite{Bogatinovski2019}. The results present the predictive performance of 26 MLC methods on 40 data-sets assessed with 18 predictive and 2 efficiency performance measures. We use the following MLC methods:
\begin{itemize}
   \item \emph{Problem transformation methods}: BR, CC, CLR, EBR, ECC, ELP, EPS, HOMER, LP, MBR, RSLP, RAkEL, PSt, TREMLC, CDE, AdaBoost, CDN, SM, CLEMS
   \item \emph{Algorithm adaptation methods}: BPNN, MLARM, MLTSVM, MLkNN, PCT, RFPCT, DBN
\end{itemize}
\noindent Details on the methods, the datasets and the evaluation measures are provided in the Supplementary material. For clarity of presentation, we used up to 10 evaluation measures in the corresponding evaluation scenarios.

The experimental pipeline contains three stages and adheres to literature recognized guidelines for conducting large-scale experimental studies~\cite{Caruana2006}.
The first stage performs a hyperparameter optimization of the methods where needed. Typically, these MLC methods have a base learner (binary or multi-class classification method). The number of hyper-parameters to be tuned for the methods can be significant. For practical feasibility, two constraints are imposed. First, if the data sets have more than 1000 training examples, they are sub-sampled to 1000 samples using an iterative stratified sampling technique~\cite{Sechidis2011}. The second constraint is having a time budget of 12 hours to find the best hyperparameter configuration via optimization of the hamming loss performance criteria. The selection of the best hyperparameters is done on a three-fold CV using the iterative stratified sampling technique. For the hyperparameter values, the recommended ranges from the literature are utilized. As base learners, J48 and SVMs are evaluated. 

The second stage involves learning a model using the complete training set. After the model is built on the training set, it is used to make predictions on the test set. 

The third stage calculates all of the evaluation measures using the stored predictions for the examples from the test set. To create bi-partitions of the labels by imposing a threshold on the predictions (where needed), the PCut~\cite{Read2010} method is used. PCut selects the threshold minimizing the difference between the label cardinally of the training and the (predictions of the) test set.

\section{Assessing the Potential of the Meta Features}

\begin{figure*}[!t]
\includegraphics[width=1.0\textwidth]{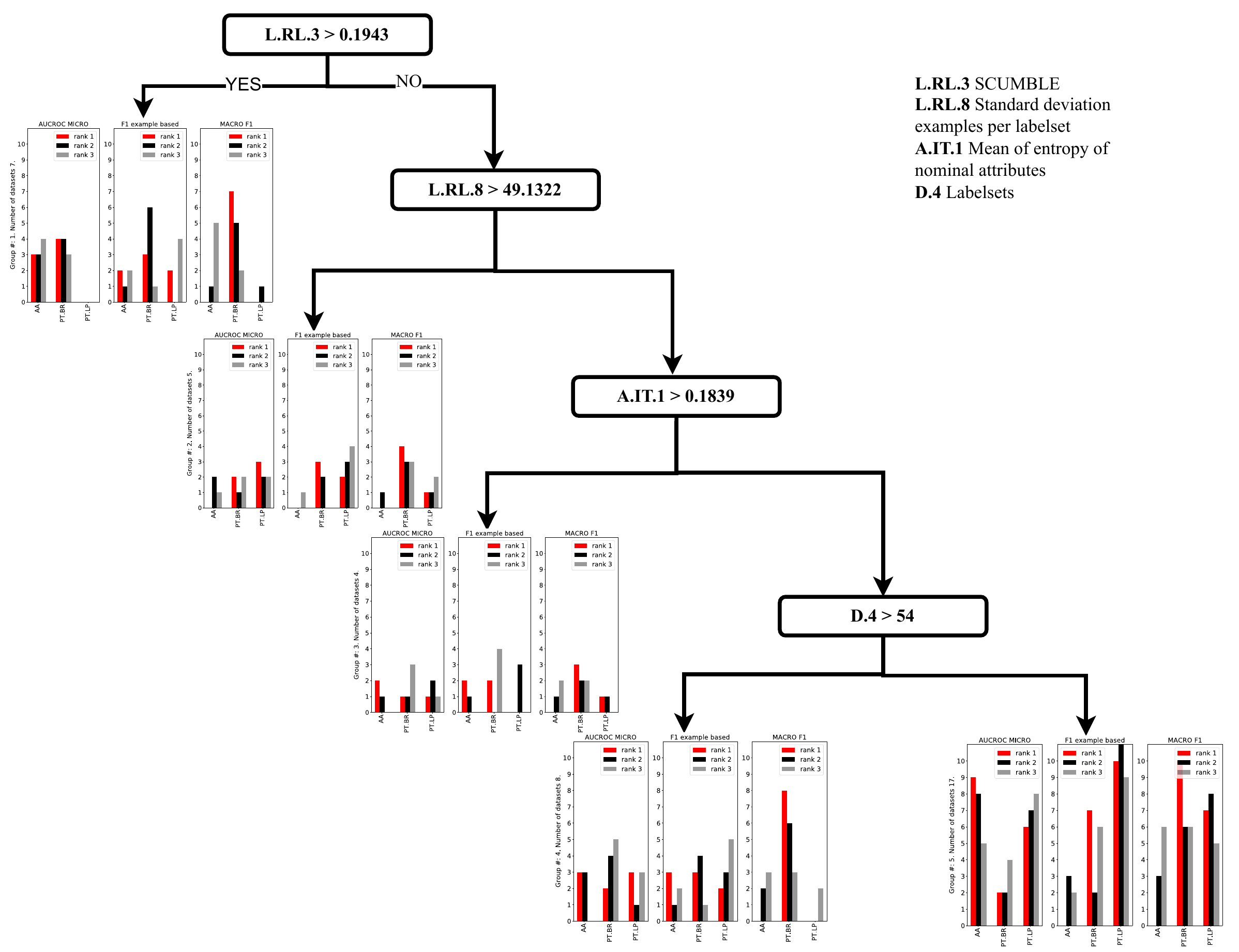}
\caption{A divisive clustering tree of the 40 data sets in the Euclidean space of the meta features. The leaf nodes correspond to clusters  and illustrate the  distribution of the top-performing MLC methods for the data sets in the cluster with respect to three performance measures (\texttt{AUROC.micro}, \texttt{F1.example.based} and \texttt{F1.macro}). The distributions are aggregated with respect to the MLC families of methods: 1) algorithm adaptation (AA), 2) problem transformation using binary relevance (PT.BR), and 3) problem transformation using label powersets (PT.LP).}
\label{sec4_1:fig:2}
\end{figure*}

\sloppy
\figurename~\ref{sec4_1:fig:2} depicts a divisive clustering tree. It was learned using only the values of the meta features, in an unsupervised manner. The extracted clusters were then represented with the performances of the methods grouped by the family of methods and described with the most influential and divisive meta features. In other words, the leaf nodes correspond to the clusters of data sets closer to each other in the meta-features space, while the inner nodes provide meta-feature explanations of the clusters' membership. In each leaf node, we observe the relative performance of the MLC paradigms of AA and PT.LP, and PT.BR. The bar diagrams depict the number of data sets in the corresponding clusters for which a method from the specific MLC paradigm is among the top-3 performing methods. We assess the methods' performance with three evaluation measures: \texttt{F1.example-based}, \texttt{F1.macro} and \texttt{AUROC.micro}. We next discuss this tree in more detail. 

The most decisive meta feature in the root node is the SCUMBLE index (\texttt{L.RL.3}). It quantifies the variation of the imbalance among the (frequent and infrequent) labels in the examples, and it is highly correlated to the meta feature \texttt{Proportion of Distinct Labelsets}. The left child of the root node corresponds to the cluster of seven data sets with high SCUMBLE index. For those data sets, the top-performing family of MLC methods is PT.BR. The ability of PT.BR methods to address the data set with high inter-class imbalance is related to the fact that they treat each label separately, unlike the PT.LP or AA methods that consider example labels jointly and thus fail to handle high imbalance. 

Down the tree, the second cluster includes five data sets with high \texttt{Standard deviation of examples per labelsets} (\texttt{L.RL.8}), another meta feature concerning the relationships among labels, which indicates the ratio of outliers among the label sets. The MLC paradigms show more uniform performance on the data sets in this cluster. PT.LP works better, owning to the Pruned Set method that discards infrequent label sets (i.e., outliers).

Note that in the first two clusters above, AA methods show slightly worse performance when compared to the PT methods. For the following two clusters down the tree, this slight difference diminishes to almost none. The third cluster includes four data sets with high \texttt{Mean of entropy of nominal attributes}, \texttt{A.IT.1} (with low SCUMBLE and a low ratio of label-set outliers), while the fourth cluster is characterized with seven data sets with a high number of \texttt{Labelsets}, \texttt{D.4}, i.e., a high number of distinct label sets. The discriminative property between the latter two clusters and the first one (where the PT.BR paradigm dominates), is the average number of examples per label set. In the first cluster, there is one example per label set on average, while for the fourth cluster this average increases to six examples per label. Thus, AA methods work well on data sets with a larger number of well-balanced label sets.
s

Finally, the last cluster at the right-hand side of the tree includes 17 data sets with a relatively small number of label sets and well-balanced distribution of labels. The methods that can directly exploit the target label sets or are simultaneously considering them, perform well. More specifically, PT.LP are top performers concerning the \texttt{F1.example-based} measure, PT.LP together with AA methods are top performers concerning the \texttt{AUROC.micro} measure, while both problem transforming paradigms perform well in terms of \texttt{F1.macro}. It is interesting to note that the majority of the data sets in this cluster are associated with the domain of bioinformatics.


In summary, when it comes to assessing the potential of the meta features for explaining the performance of the MLC methods, our results show that PT.LP performs very well across all the performance measures for data sets with a small number of label sets and a high number of examples per label set. On the other hand, PT.BR perform well in data sets with an extreme imbalance among label sets and a small number of examples per label set. AA methods perform well when there are quite a few samples per label sets and there is absence of extreme imbalances within the label sets. This clear cut between the "domains of expertise" of the different MLC paradigms confirm that our meta features hold promise for explaining the performance of MLC methods, which we explore in the next section. The cluster membership of the data sets is decided mostly upon the meta features assessing different aspects of the label space (two out of four meta features are from this category).

\section{Relating Predictive Performance with Meta Features}
\begin{table}
\caption{Mean absolute error (MAE) of the multi-target meta models for predicting the performance of RFPCT, EBRJ48 and RFDTBR (three table columns) wrt different evaluation measures (table rows). The last two columns report the MSE of the model averaged over the three methods and the baseline model that predicts the mean values of the targets in the training data set.}
\label{tab:scorePredictionMTR}
\centering
\resizebox{0.8\textwidth}{!}{%
\begin{tabular}{|l|r|r|r|r|r|}\hline
Evaluation measure          & RFPCT (PCT) & RFDTBR (PCT) & EBRJ48 (PCT)  & Average (PCT)  & Baseline \\\hline
AUROC.micro       & $0.066$ & $0.066$ & $0.073$ & $0.068$   & $0.057$ \\ \hline
F1.example-based   & $0.126$ & $0.134$ & $0.149$ & $0.135$   & $0.179$ \\ \hline
Hamming loss       & $0.057$ & $0.058$ & $0.063$ & $0.059$   & $0.066$ \\ \hline
F1.macro           & $0.164$ & $0.172$ & $0.189$ & $0.174$   & $0.221$ \\ \hline
F1.micro           & $0.143$ & $0.142$ & $0.157$ & $0.147$   & $0.165$ \\ \hline
\end{tabular}
}
\end{table}

In this section, we build meta models for predicting the performance of selected MLC methods on a new unseen data set. As a descriptive space, we use all the 50 meta features. The target changes depending on the learning scenario. First, we predict the absolute performance of the selected methods (MTR problem). Second, we predict the relative performance for the methods with respect to one another (ST problem). Finally, we consider a single target regression scenario where we investigate the efficiency of the learning methods with respect to the training and prediction (test) time. 


Following the results reported in~\cite{Bogatinovski2019}, we focus on the performance of three MLC methods: RFPCT~\cite{Kocev:Journal:2013,Madjarov2012}, RFDTBR~\cite{Tsoumakas2007} and EBRJ48~\cite{Read2010} that cover the two MLC families of methods: AA and PT. To obtain compact predictive models, we employ multi-target regression (MTR) trees that predict the performance of the three methods simultaneously. For tractability of the results we learn five MTR trees corresponding to the evaluation measures of \texttt{AUROC.micro}, \texttt{F1.example-based}, \texttt{hamming loss}, \texttt{F1.macro} and \texttt{F1.micro}. The considered measures cover at least one measure in each group of performance evaluation criteria concerning predicting bipartitions.  For learning the MTR trees, we tune the hyperparameter F-test from the set of candidate values $\{0.001, 0.01, 0.05, 0.1, 0.125\}$ . We select the value that leads to minimal mean absolute error measured following a leave-one-dataset-out procedure.

\begin{figure*}[!t]
\centering

\subfloat[F1.example-based (MTR)]{\includegraphics[width=0.5\textwidth]{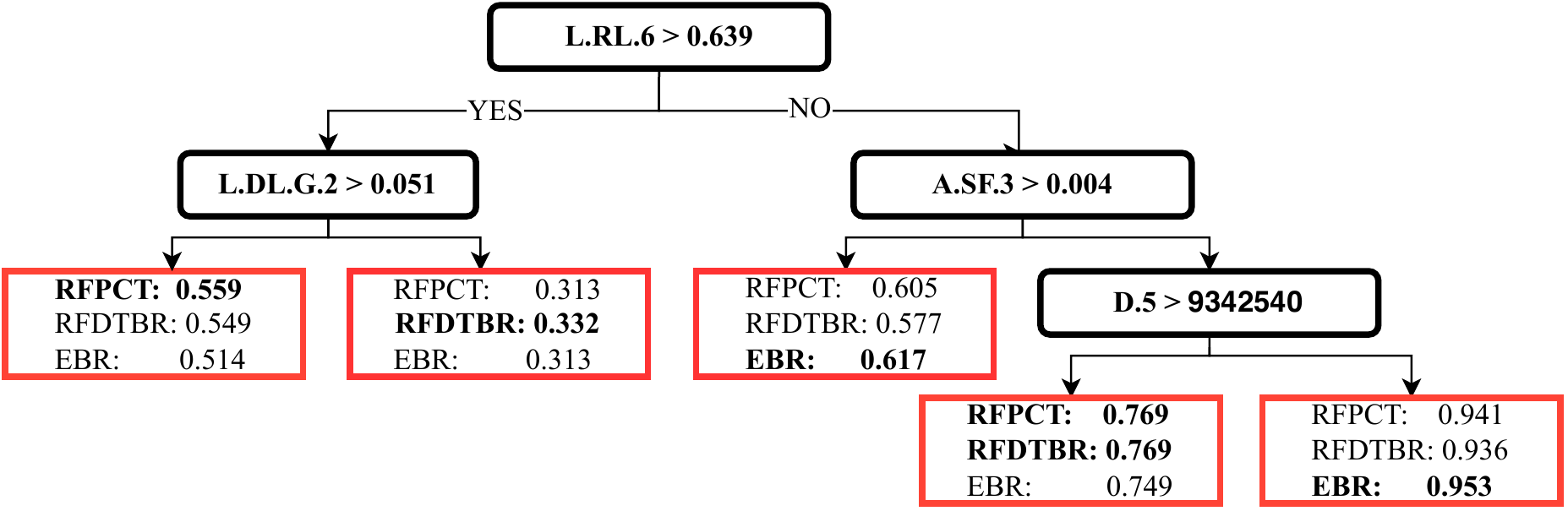}%
\label{sec4_3:fig:10}}
\hfil
\subfloat[F1.example-based (ST)]{\includegraphics[width=0.5\textwidth]{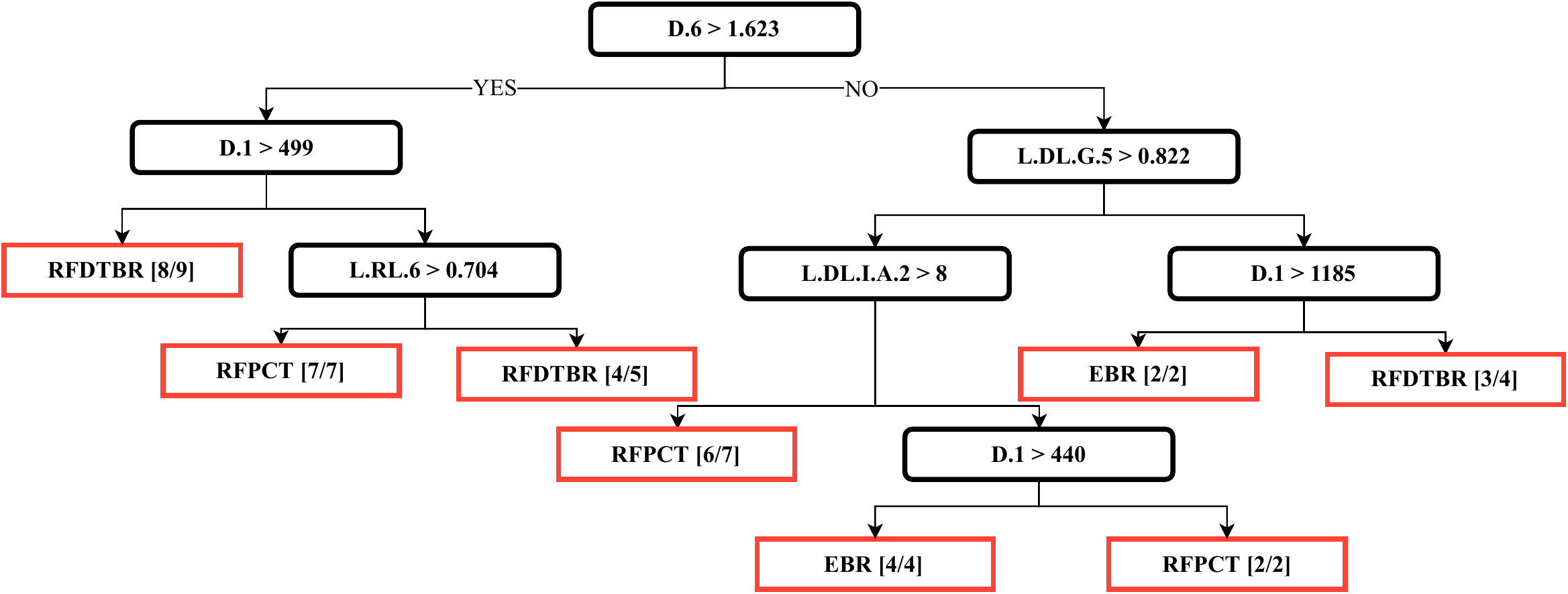}%
\label{sec4_3:fig:12}}


\subfloat[Hamming Loss (MTR)]{\includegraphics[width=0.5\textwidth]{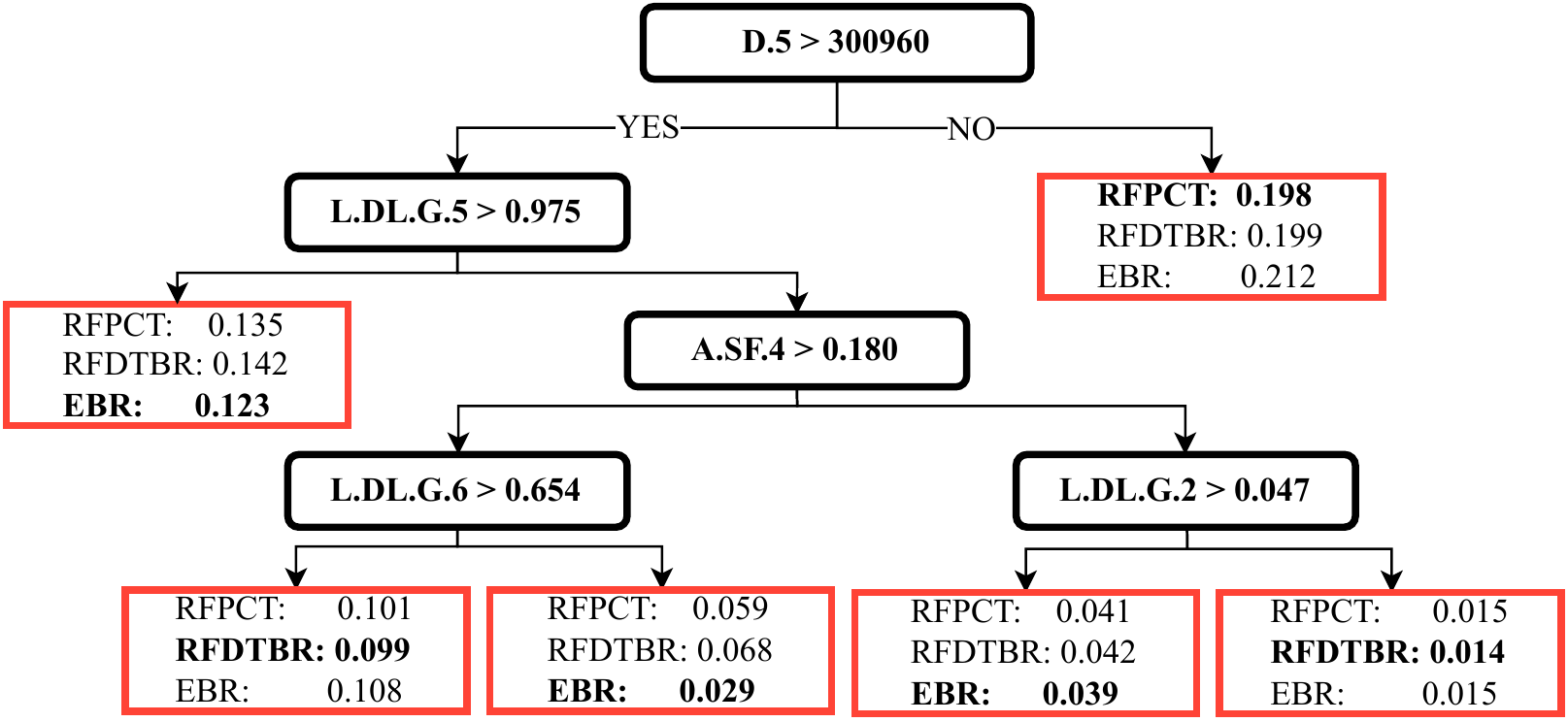}%
\label{sec4_3:fig:9}}
\hfil
\subfloat[Hamming Loss (ST)]{\includegraphics[width=0.5\textwidth]{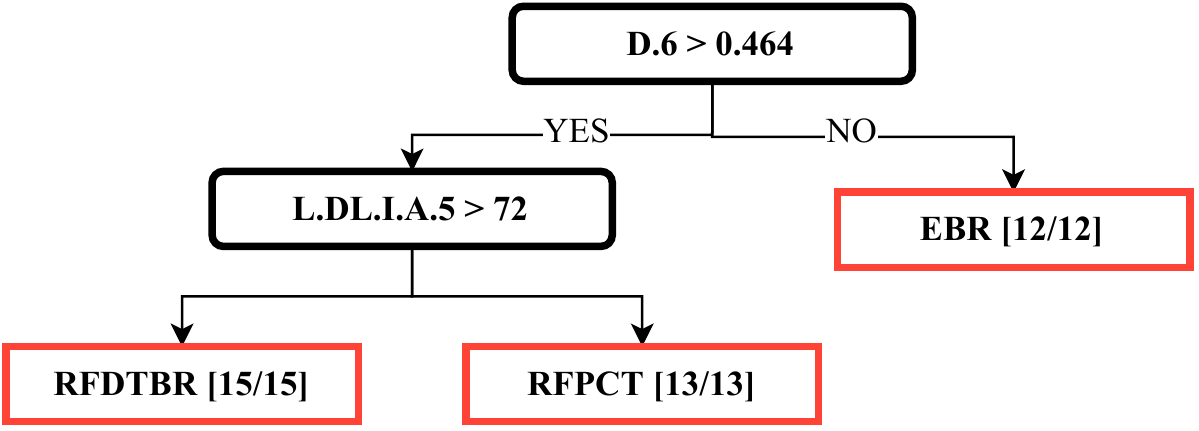}%
\label{sec4_3:fig:14}}

\caption{Multi-target regression trees for predicting the performance of the MLC methods (left-hand side) and classification trees for selecting the best-performing MLC method (right-hand side). For better reading we provide the corresponding abbreviations \textit{D.1} Attributes, \textit{D.5} LxIxF, \textit{D.15} Proportion of unique label combination, \textit{D.6} Ratio of number of Instances to number of Attributes,  \textit{A.SF.3} Mean of skewness of numeric attributes, \textit{A.SF.4} Mean of mean of numeric attributes, \textit{L.RL.6} Ratio of examples up to 50 examples, \textit{L.RL.11} Number of unconditionally dependent label pairs by chi-square test, \textit{L.DL.G.2} Density,  \textit{L.DL.G.6} Kurtosis Cardinality,  \textit{L.DL.I.A.2} Maximal of Imbalance Ratio intra-class, \textit{L.DL.I.A.5} Maximal Imblanace Ratio per labelset.}.
\label{fig:MultiClass}
\end{figure*}

\tablename~\ref{tab:scorePredictionMTR} summarizes the performance of the meta models assessed with different evaluation measures. For F1 (micro, macro and example-based) and Hamming loss, the MTR model outperforms the baseline by 15 to 30\%. The improvements over the baseline are largest for the three F1-based evaluation measures, so we depict the corresponding MTR trees in \figurename~\ref{fig:MultiClass} and analyze the meta knowledge they include. For the \texttt{AUROC.micro} evaluation measure, we fail to obtain a model that would outperform the baseline that predicts the mean values of the evaluation function for the examples in the training data set. Nevertheless, the MTR model explains the variance, preserving the interpretability property over the baseline. We depict and discuss the MTR models for the F1.example-based and hamming loss (as measure being optimized). The description of the remaining 3 models is given in the Supplementary. 

\figurename~\ref{sec4_3:fig:10} depicts the MTR tree for predicting the \texttt{F1.example-based} evaluation measure. The root node splits the data sets on the ratio of the number of label-sets with up to two examples (\texttt{L.RL.6}). For the data sets with the ratio higher than 64 \% (left branch of the tree), the three methods lead to the performance below 0.6, while for the other data sets (right branch) the performance is mostly above 0.6. In the left branch, the data sets are further split upon the label density (average ratio of labels assigned to an example, \texttt{L.DL.G.2}). The worst MLC method performance, below 0.35, is observed for data sets with low label density. For data sets with higher label density, performance grows notably to at least 0.5. Down the right branch of the tree, the data sets are split upon the value of \texttt{A.SF.3}, the mean skewness of the numerical attributes. High mean skewness leads to lower performance (below 0.65). The data sets with higher mean skewness are finally split upon their complexity (\texttt{D.5}, the product of the numbers of instances, attributes and labels), where lower complexity leads to highest performance above 0.9 (right-most leaf node of the tree). Note that two top-level splits in the tree include meta features assessing the distribution of the labels, while one split is based on the data set complexity, which depends on the number of labels as well.

\figurename~\ref{sec4_3:fig:9} depicts the results for the comparison of the three methods on the \texttt{hamming loss} evaluation measure. The root node suggests splitting of the datasets in terms of the complexity of the MLC datasets given as a product of the number of features, instances and labels. For the datasets with smaller complexity, the RFPCT method is slightly preferred over the other methods. For the remaining datasets, two extremes appear. If the entropy of the labels is above 0.975, there is a stronger preference of EBRJ48 in comparison to the RFDTBR and RFPCT as well. On the other side for the datasets with larger "label density" ($L.DL.G.2 \geq 0.047$) and small "kurtosis cardinality" ($L.DL.G.6 \leq 0.654$) EBR is preferred in comparison to the other two methods.  

\begin{table}
\centering
\caption{Accuracy of the classification trees for predicting the best-performing MLC method compared to the ratio of the meta examples in the majority class (baseline).}
\label{multiclassRes}
\resizebox{0.5\textwidth}{!}{%
\begin{tabular}{|l|r|r|}
\hline
Evaluation measure        & PCT multi-class & Baseline \\ \hline
AUCROC.micro     & $0.525 $        & $0.550$  \\ \hline
F1.example-based & $0.525 $        & $0.425$   \\ \hline
Hamming loss     & $0.550 $        & $0.450$   \\ \hline
F1.macro         & $0.475 $        & $0.475$   \\ \hline
F1.micro         & $0.625 $        & $0.525$   \\ \hline
\end{tabular}
}
\end{table}

\tablename~\ref{multiclassRes} summarizes the performance of the single target (ST) meta models assessed with five different performance scores. For F1 (micro, macro and example-based) and Hamming loss, the multi-class PCT outperforms the baseline by up to 22\%. The improvements over the baseline are largest for the three F1-example-based evaluation measures. Therefore, we depict the corresponding models in \figurename~\ref{fig:MultiClass} and analyze the meta knowledge it includes. The results for the remaining performance criteria are given in the Supplementary material. For learning the ST trees, we tune the hyperparameter F-test from the set of candidate values $\{0.001, 0.01, 0.05, 0.1, 0.125\}$ . We select the value that leads to minimal mean absolute error measured following a leave-one-dataset-out procedure.

\figurename~\ref{sec4_3:fig:12} depicts the classification tree for predicting the best-performing MLC method with respect to the \texttt{F1.example-based} evaluation measure. Two splits at the top of the tree are based on the size of the data set measured in terms of the ratio of examples over the number of features (\texttt{D.6}), and the number of features (\texttt{D.1}). The tree suggests that for large data sets with many features, the RFDTBR method performs the best. For data sets with a large ratio of examples over features, less than 500 features, and the ratio of label-sets with up to two examples (\texttt{L.RL.6}) higher than 70\%, RFPCT is the best-performing method. In the right-hand branch of the tree, if the high maximal entropy of labels (\texttt{L.DL.G.5}) is greater than 0.832, then RFPCT prevails. The node with four data sets with less than 1.6  for the ratio of the number of instances to the number of features, high maximal entropy of labels (\texttt{L.DL.G.5} greater than 0.832), maximal imbalances of the intra-class between the labels (\texttt{L.DL.I.A.2} less than 8), and more than 440 features (\texttt{D.1}), is the node where EBRJ48 outperforms RFPCT. If the maximal entropy of the labels is smaller than 0.832 then the problem transformation methods prevail. The EBRJ48 outperforms RFDTBR if the number of features (\texttt{D.1}) is greater than 1185.

\figurename~\ref{sec4_3:fig:14} depicts the classification tree for predicting the best-performing MLC method with respect to \texttt{hamming loss}. The induced tree provide clear preferences for the three methods in different settings. More specifically, the datasets with a small ratio of attributes to instances ($D.6\leq0.464$) prefer the EBRJ48 method. Comparing to other datasets with large imbalances of the labels, the RFDTBR should be performed as opposed to RFPCT.

\begin{figure*}[!t]
\centering
\subfloat[Meta model efficiency (training). ]{\includegraphics[width=0.5\textwidth]{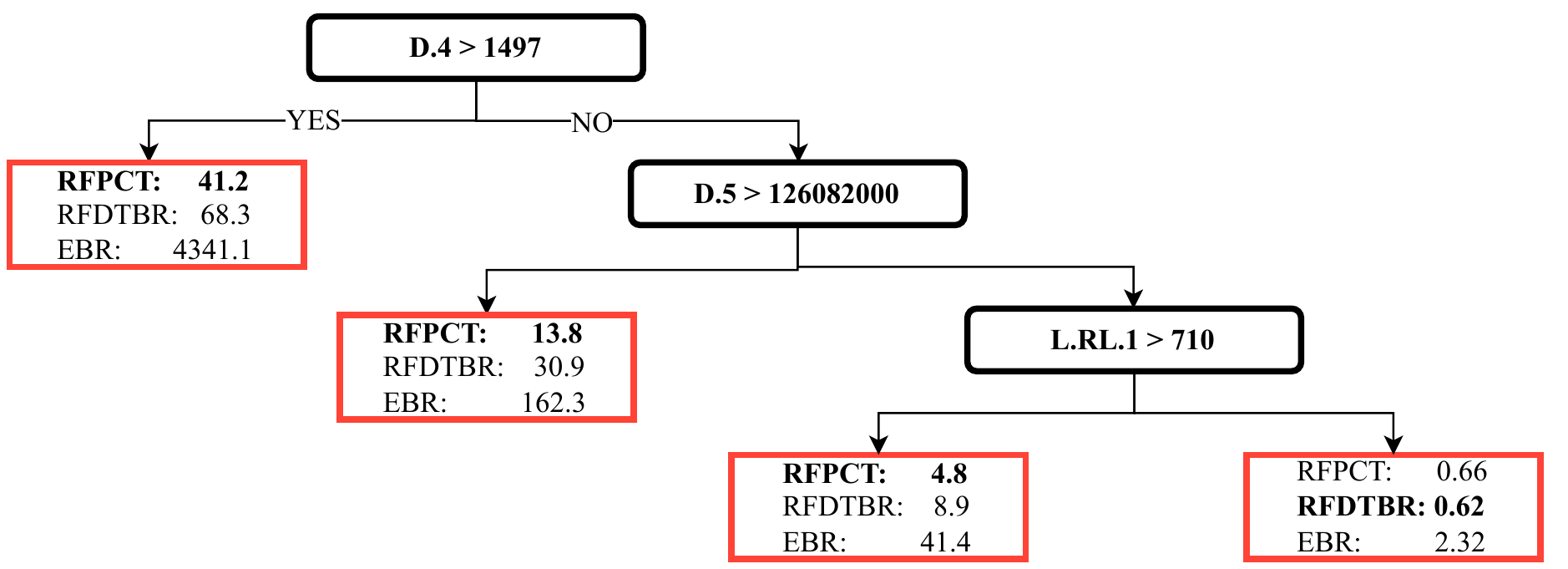}%
\label{sec4_3:fig:15}}
\hfil
\subfloat[Meta model efficiency (test).]{\includegraphics[width=0.5\textwidth]{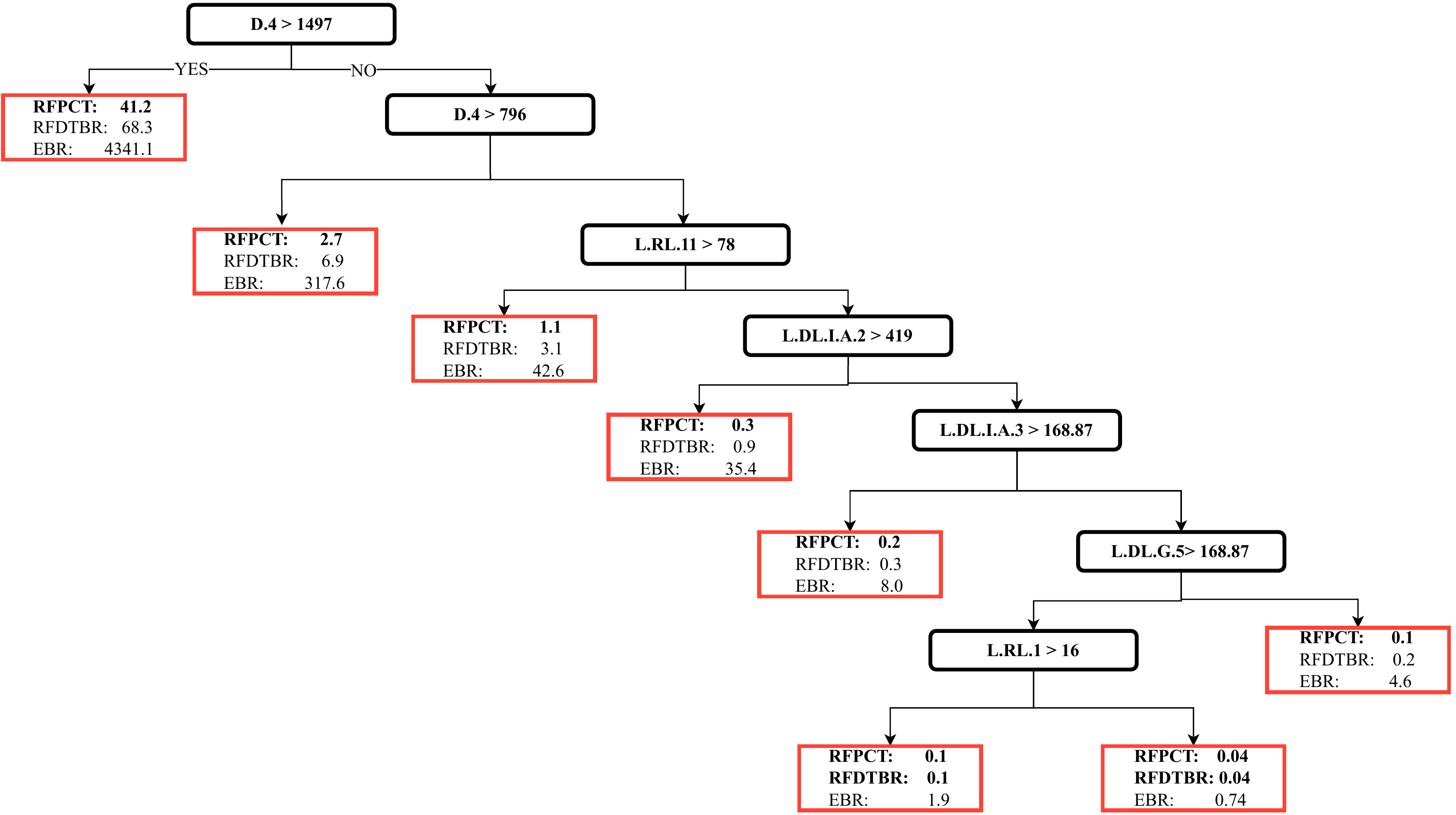}%
\label{sec4_3:fig:16}}
\caption{Classification trees predicting the MLC paradigm that yields the best predictive performance.}
\label{fig:times}
\end{figure*}

\figurename~\ref{fig:times} shows the preference for the different methods concerning their efficiency. The calculated time includes the time needed to obtain the predictions with the PCut method. For the training and the test time, the meta features characterizing the dataset size, appear at the top of the tree. An interesting observation because in line with the intuition that the number of instances, features and labels influence is that the time needed to build models and infer predictions. The remaining attributes refer to the features describing the label space. For training time, the number of unique label sets (for some datasets this is related to the number of instances), as well as the complexity of the datasets given as a product of the number of features, samples and labels, tend to separate the tree, outlining four groups of datasets. For three of them, the RFPCT method is faster as opposed to the other two methods. Since the underlying heuristic of RFPCT is calculated for all the labels simultaneously, the trees in the ensemble are smaller, therefore faster to built. When the datasets are smaller, with fewer labels (the last node in the tree), this advantage diminishes, and RFDTBR is equally efficient. EBRJ48 is applying the BR procedure multiple times, therefore, it constantly requires several iterations over all labels. Therefore, EBRJ48 is persistent in under-performing in terms of efficiency.  \figurename~\ref{sec4_3:fig:16} provide conclusions that are in line with the conclusions from the training procedure.

\section{Investigating the need for parameter tuning}

In the last series of meta experiments, we want to assess and analyze the performance gain obtained by tuning the hyperparameters of the MLC methods. We start with analyzing the ratio of successful parameter-tuning experiments, then we analyze the sensitivity of MLC methods and paradigms to varying hyperparameter settings, and finally, we explore the main issue of whether and when to tune.

\subsection{Ratio of Successful Experiments}

When performing the base-level experiments with the MLC methods on the selected data sets, for each pair of a method and a data set, we performed many experiments corresponding to the number of the candidate parameter settings. To that end, we introduce RSED as a measure for the success/finishing rate of the experiments. RSED is calculated as the ratio of successful experiments per data set and method. We observe two averages of the RSED: methodRSED is the marginal average over data sets, while dsRSED is the marginal average over methods.  dsRSED is high for smaller data sets: for 60~\% of the data sets, more than 67~\% of the experiments were finished. Only in about 28~\% of the data sets, the success ratio was below 30~\%. 

\begin{figure*}
    \includegraphics[width=1.0\textwidth]{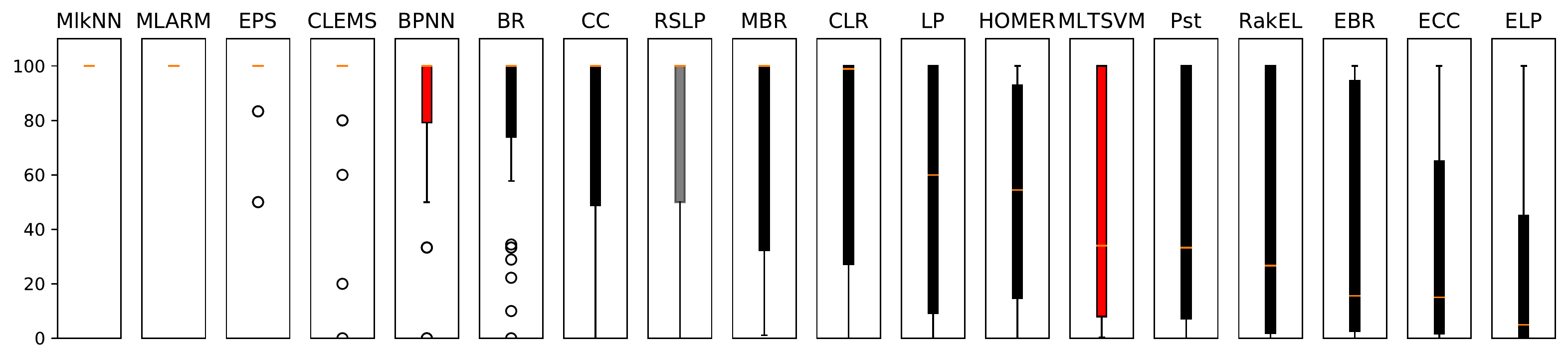}
    \caption{Box plots of the ratio of successful experiments per method (methodRSED) for 18 MLC methods. The methods are ordered with respect to the decreasing average RSED. Black denotes the problem transformation, while the red the algorithm adaptation methods.}
    \label{sec431}    
\end{figure*}

\figurename~\ref{sec431} provides further insight into the methodRSED. For the AA methods, the RSED is larger compared to the PT methods. This is because AA methods tend to build a single model (in the case of ensembles, the base MLC method is AA that builds one method). The PT methods are highly affected by the number of labels. Since they are training multiple single target models as base learners per label or label sets, they need several iterations over the descriptive space. For example, ELP is a method that utilizes SVM as base learner~\cite{Moyano2018}. It builds multiple LP models where the solution of the induced multi-class problem is done using \texttt{One-Vs-One} approach. For example, if a dataset has 20 labels, there are million possible classes to choose from. Similar observations can be made for RAkEL, HOMER, ECC and EBR~\cite{Read2010}. Since HOMER and RAkEL are facing a balanced distribution of the labels, their RSED is larger than the ones for ECC, EBR and ELP. Note also that some of the implementations of the methods require that the learned base-level classifiers are stored in memory during learning, which might influence the ratio of successful experiments per method.

\subsection{Sensitivity to Hyperparameter Tuning}

\begin{figure*}[!t]
    \includegraphics[width=1.0\textwidth]{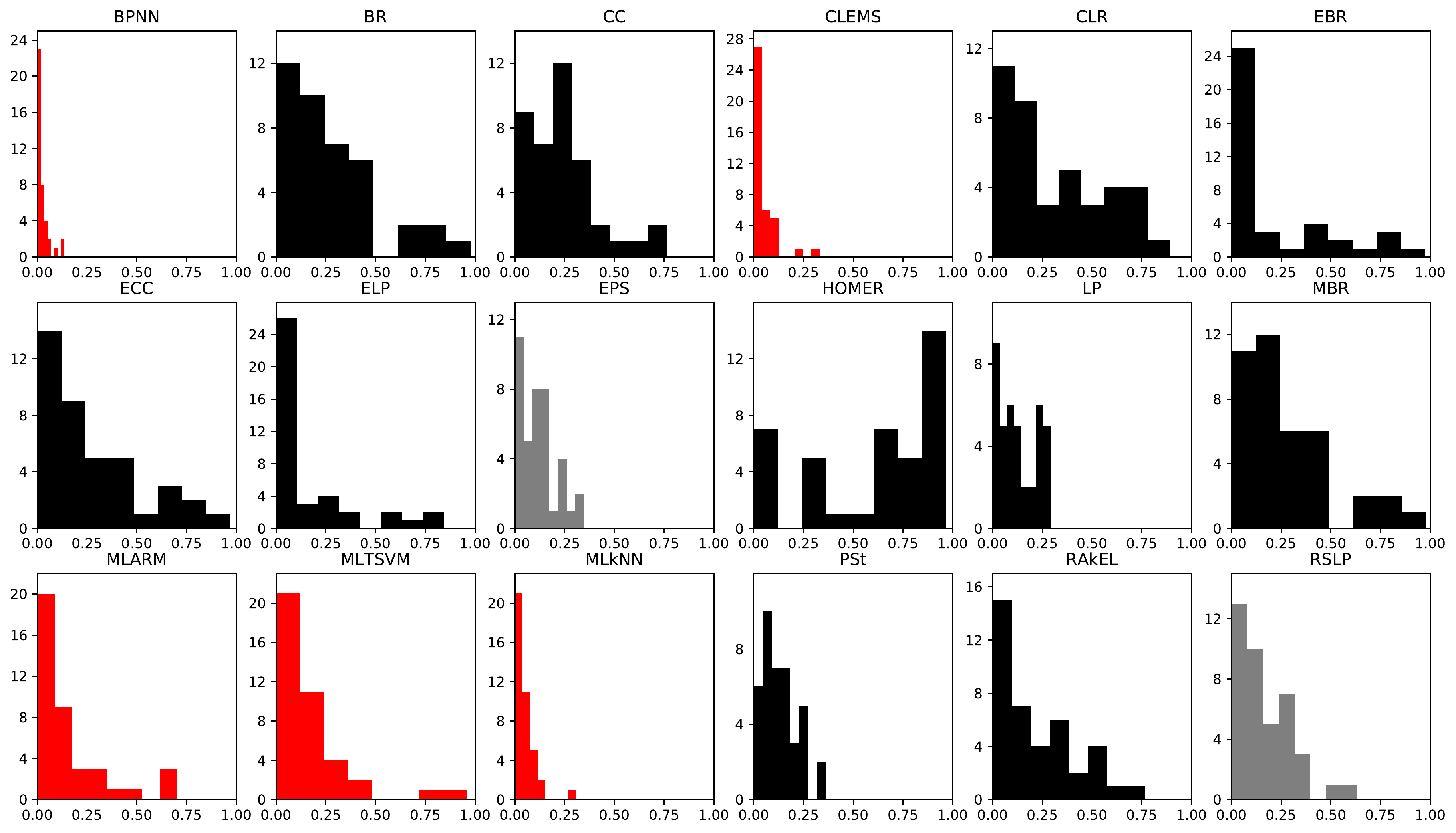}
    \caption{Histograms of the relative improvement of the hamming loss gained through hyperparameter tuning over the data sets for each of the MLC methods. Black denotes the problem transformation, while the red the algorithm adaptation methods.}\label{sec4_2:fig:2}
\end{figure*}

We next explore the sensitivity of the MLC methods concerning the varying values of the hyperparameters. For a given method and a data set, we observe the improvement of the hamming loss gained through hyperparameter tuning with respect to default values. When the default hyperparameter setting leads to a best hamming loss, the relative improvement of 0 is observed, as opposed to the extreme situation when the tuned hyperparameter setting has a hamming loss of 0 and therefore leads to the maximal relative improvement of 1. \figurename~\ref{sec4_2:fig:2} depicts the distribution of the relative improvement over the data sets for each MLC method: each graph depicts the distribution of the number of data sets with a certain value of the relative improvement. Skewed distribution with peaks clustered on the left-hand side (relative improvements close to 0) indicates that it is hard to improve upon the performance of the default hyperparameter setting of the method. \figurename~\ref{sec4_2:fig:2} further show that most of the methods (with a notable exception of HOMER) have positively skewed distributions, showing that, in most cases, MLC methods are not very sensitive to the hyperparameter settings, therefore not much gain can be expected from hyperparameter tuning. Furthermore, when using AA methods, we do not have any significant improvements for hyperparameter tuning.

The problem transformation methods are characterised by a larger variety of relative improvement. 
When observing the results for the methods that use SVM as the base learner, a large deviation in the rankings is observed. Due to this high volatility, it is not clear if the change in rankings is due to the hyperparameters of the MLC methods or the hyperparameters of the base learner. As suggested in~\cite{vanRijn2018, Mantovani2015}, tuning of the \textit{cost} and \textit{gamma} hyperparameters for SVM with Gaussian kernel is desirable in the case of binary classification. If the other parameters of the method have greater importance, it is expected that the influences of both the parameters of the base learner (in this case SVM) and the remaining parameters of the MLC method to experience a synergistic effect. The RSLP and EPS methods are using J48. For these methods, the improvement is negligible (for most of the datasets is between 5-10~\%). Recalling to the "static" behaviour of J48~\cite{Mantovani2018} for binary classification concerning the tuning of the \texttt{cost complexity pruning} hyperparameter, we can conclude that for both EPS and RSLP, the default parameters are reasonably set. Their tuning can provide benefits, however, oftentimes marginal.

\subsection{To tune or not to tune?}
We design the meta analysis to consider two categories of MLC methods: \emph{reliable-defaults} and  \emph{hyper-tuned} methods. The former group methods includes RFPCT, AdaBoost~\cite{Schapire2000}, TREMLC\cite{Tsoumakas2010b}, EBR with J48 , ECC with J48~\cite{Read2010}, BR with RFDT and CDE~\cite{Chekina2010} that provide good, hard-to-improve-upon default settings of the hyperparameters. The latter group of methods includes MLC methods for which substantial performance improvement can be gained with parameter tuning. Therefore, a meta-learning question emerges for a given data set: whether to use a strong-baseline method or a hyper-tuned one.  

\begin{figure*}
    \centering
    \includegraphics[scale=0.43]{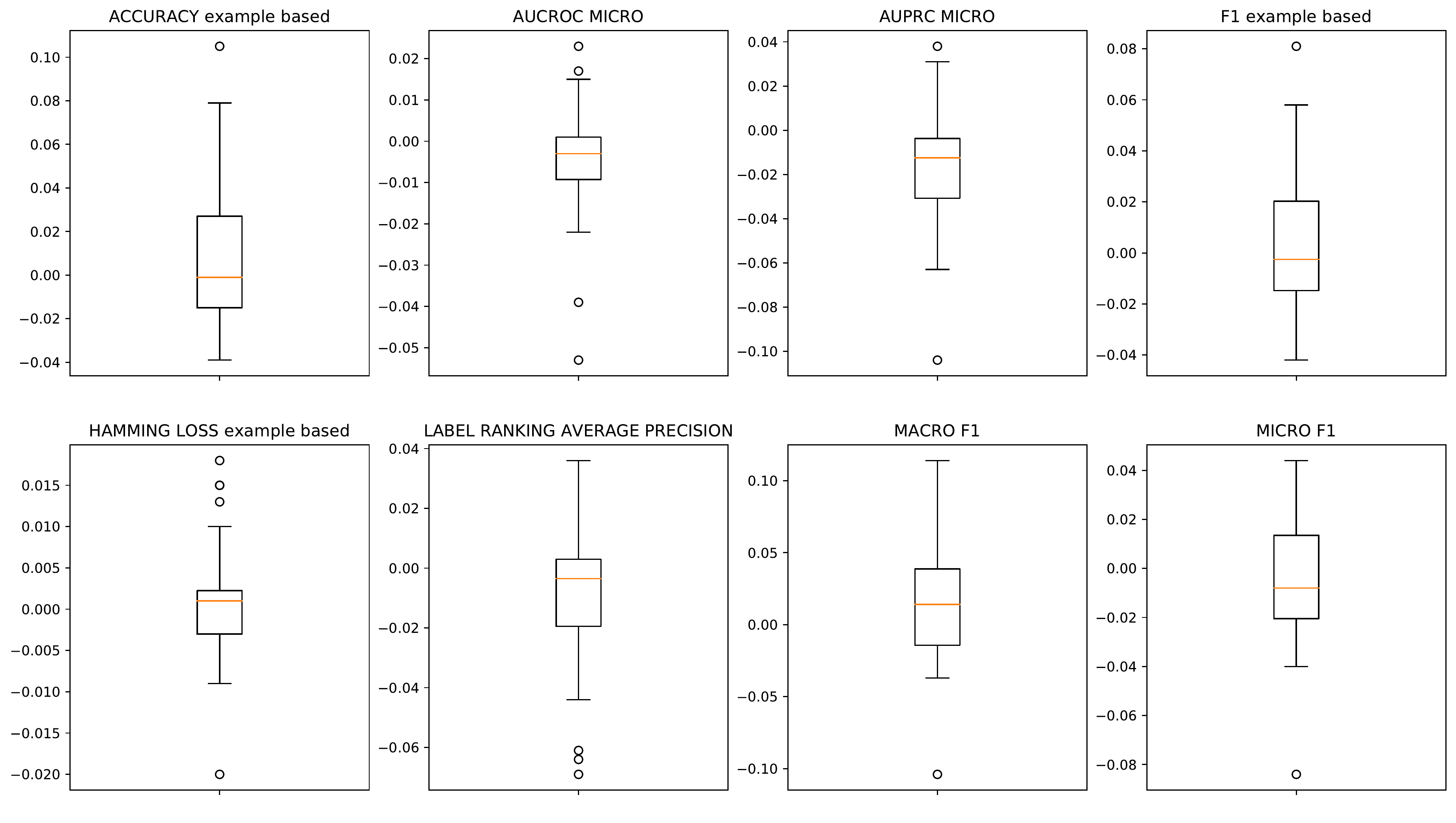}
    \caption{Box plots of the differences between the out-of-sample (test) performances of the model learned with a reliable-defaults methods and the one learned with hyper-tuned methods for each of the evaluation measures.}
    \label{sec4_2:fig:3}  
\end{figure*}

\figurename~\ref{sec4_2:fig:3} depicts the results of the comparison. It depicts the relative difference between the best method from the \emph{hyper-tuned} group and the best method from the \emph{reliable-defaults} group on the corresponding evaluation measure marginalized over all datasets. On the eight evaluation measures can be seen that according to \texttt{F1.macro} and \texttt{hamming loss} the optimization of the hyperparameters is beneficial. 

\begin{figure}[!h]
    \centering
    \includegraphics[width=0.8\textwidth]{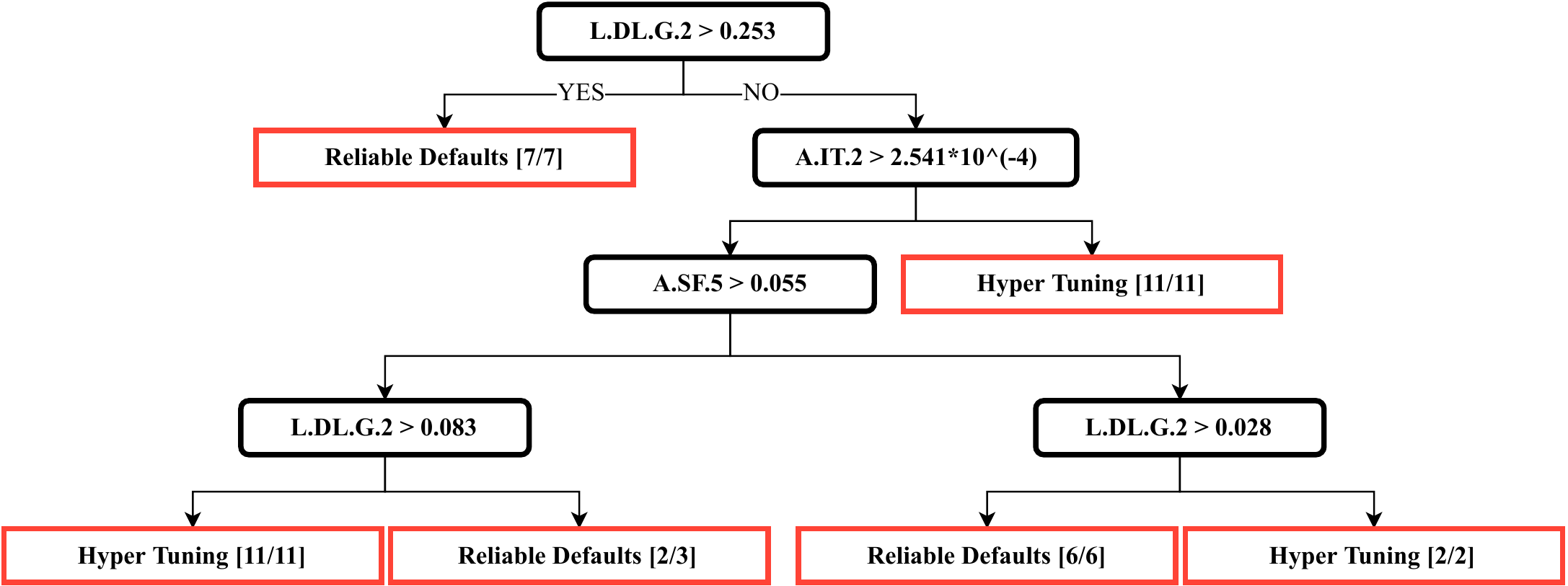}
    \caption{A classification tree deciding upon whether to use a reliable-defaults or a hyper-tuned MLC method for a given data set. The tree has an accuracy of 0.68, AUROC of 0.66 and AUPRC of 0.72.}
    \label{sec4_3:fig:4}  
\end{figure}

Finally, we formulate the selection between the reliable-defaults and hyper-tuned methods as a classification task and learn the classification tree depicted in \figurename~\ref{sec4_3:fig:4}. The tree suggests that tuning of the hyperparameters depends predominantly on the properties of the labels and to some extent on the features of the data sets. For the data sets with higher density (\texttt{L.DL.G.2}) - high frequency of the labels (above 0.253), a good performing method can be inferred straight from \emph{reliable-defaults} methods. Similarly, if the separability of the classes is greater - larger average gain ratio (\texttt{A.IT.2}, greater than 0.00025) for a certain range of value for the density (between 0.083 and 0.028) the \emph{reliable-defaults} are recommended. Otherwise, hyper tuning is preferred.

In summary, hyperparameter optimization can help to improve performance. The degree of improvement depends on the evaluation measure. The extent to which the improvement is significant in general does not always justify the utilization of the resources, both time and computation.

\section{Enriching the MLC knowldege corpus}
In the following, we highlight our major contributions to the MLC knowledge corpus.
\emph{(1) What meta features should we measure on a given data set and how do they influence the selection of MLC methods?}

We propose a set of four easy to implement and fast to compute meta features that can serve the practitioners as reliable features to make an informed decision for the most suitable family of methods to select from.  The identified features determine the "domains of expertise" of the MLC paradigms and methods. As a side contribution, we provide a novel taxonomy of the meta features for the MLC task.

It complements the existing literature that relies on an expert selection of the meta features and expert imposed thresholds for their values, as guidelines. It enhances the set of informative features by providing bigger emphasis on the meta features depicting a greater variation of the imbalance among the  (frequent and infrequent)  labels (\texttt{L.RL.3}), the variation of the number of samples for the label sets (\texttt{L.RL.8}), mean entropy of the nominal features (\texttt{A.IT.1}) and the number of labels (\texttt{D.4}) as opposed to the commonly paired meta features of the ratio of dependent label set pairs (\texttt{L.RL.12}), imbalances between the labels (\texttt{L.DL.I.E.1}) and approximate complexity (\texttt{D.5})~\cite{Moyano2018}.

\emph{(2) Which MLC methods and in what particular order should we select and/or try on a given data set?}

The order of the selection of a particular method to be used depends on the performance criteria being optimized. For \texttt{F1.macro}, the selection is always between RFDTBR and EBR, with the former being preferred. However, for \texttt{F1.example-based} and \texttt{F1.micro} the meta features determine different interesting regions of dominance. The RFPCT method performs well when there are a relatively greater number of instances compared to the number of features and there is higher label density. RFDTBR and EBR perform the best in situations where there is a higher SCUMBLE (\texttt{L.RL.3}) index. Providing good scores among all the performance evaluation criteria, a practitioner should examine RFDTBR and EBRJ48 followed by RFPCT. 


Introducing the interpretive multi-target approach for meta learning, allow us to determine regions of similar values for the performance of the best-performing methods, constrained on the meta features of the data sets. As such the trees can be used for fast checking the possible expected performance score within the specified margin of error for each measure. Note that multi-target trees provide a more compact representation of the meta knowledge and are thus, more suitable for descriptive analysis than single-target trees.


\emph{(3) Should we perform hyperparameter tuning on a given data set?}

The results from our study suggest that the majority of the improvement for the hamming loss is condensed into a long-tailed positively skewed distribution. It reflects a belief that tuning of hyperparameters helps, however, the extent to which this tuning is beneficial might not always justify the utilization of the resources, both the time and computation for the MLC task. These conclusions follow a line of studies from the binary classification tasks, which suggest that it is better not to tune or to select scenarios when to tune as compared to the default "must-tune" scenario~\cite{Domingos2000,Mantovani2015,Girard,Mantovani2018}. We extend this conclusion into the MLC domain. We conclude that \emph{reliable-default} methods should be preferred in the cases when there are high frequency and good separability of the labels. Our study provides data-driven definitions for the meaning of "high frequency" and "good separability" of a data set.

\emph{Limitations of the study.} While having great practical relevance for detailed depicting the landscape of a learning task, these form of studies have inherited limitations. One of the drawbacks of making large experimental studies relates to that they are computationally expensive. For example, the time of computation for hyperparameter tuning as well as building the models on all of the 40 datasets, while optimizing single performance criteria took approximately 126720 CPU hours. 
This time is linearly dependent with respect to the number of performance criteria one is optimizing. Thus optimization over all of the performance criteria is practically infeasible with a lack of appropriate infrastructure. 


To overcome this challenge, we adopt design choices following recognized literature standards~\cite{Caruana2006}. The iterative stratified cross-validation strategy preserves the frequency of the labels. In the study, there are 19 datasets with more than 1000 samples and seven datasets with more than 5000 samples and two more than 9000. The iterative stratified sampling strategy although sub-samples the datasets, it aims to preserve the frequency of the label sets as demonstrated in~\cite{Sechidis2011}. Thereby, the potential effect of overfitting to the data is not expected to have sustainable influence over the choice of the best method. Even if it indeed happens for some datasets, the number of datasets where the effect is observed will be small, and will not hurt the drawn conclusions by large.

We selected hamming loss~\cite{Madjarov2012} for hyperparameter optimization, because it is analogous to error rate in binary classification. It is selected as evaluation criteria for best hyperparameter selection because it provides penalization for miss-classification of individual labels. Optimizing for other measures, e.g., F1, precision and recall (micro, macro and example-based) have inherited bias towards specific paradigms that are correlated with the assumptions done by the families of methods as their bias. Considering threshold independent measures, they discard methods that cannot produce rankings. Accuracy example-based evaluates just the correctly predicted labels, while the subset accuracy is blind to correct prediction of right and incorrect prediction of wrongly predicted labels. Thus, Hamming loss seems like the most fair choice for optimization. 

A second constraint when performing large scale study emerges from the included datasets. This is related to the maturity of the task - hence the number of datasets existing in the literature. The conclusions in such studies find their validity to hold in the meta-space constrained by the values of the meta-features of the included datasets. For MLC, to the best of our knowledge, this is the largest number of datasets and methods evaluated. Thus we believe that it depicts the current state of the field pointing out guidelines for both practitioners and experts to design and choose the most suitable methods for their MLC problem at hand and further expand the field of MLC as an important task in machine learning.

\section{Conclusion}
In this work, we provide a comprehensive meta-learning study for MLC. Namely, we performed the largest meta study on MLC methods to date. It uses 50 meta features to describe the 40 data sets. The study further covers 26 methods evaluated on 20 evaluation performance measures. This allows us to address a variety of practically relevant meta-learning questions. We summarize the work along the lines of three major meta analyses:

\begin{enumerate}
    \item {\emph{What is the potential of the meta features to describe the space of MLC data sets?}\\
The MLC meta features paint a very interesting landscape of the MLC datasets, identifying features that determine the "domains of expertise" of the  MLC  families of methods, as illustrated with the clustering tree. The proposed taxonomy facilitates an easier understanding of the knowledge from the meta models. The most prominent meta features that best sketch the landscape are the ones assessing different aspects of the label space.
    }
    \item {\emph{Whether and how the meta features are related to the predictive performance of the MLC methods?}\\
We present several meta models (i.e., PCTs) that relate the meta features with the predictive performance of three state-of-the-art MLC methods. These models can be easily used to assess the expected performance of a novel data set. Furthermore, the models show that the most interesting meta features describe the label space, and the meta features describing the relationships among the labels tend to occur a bit more often than the features describing the distributions between and within the individual labels.
We also present meta models for predicting the best performing MLC method. The learned meta models exploit meta features mainly assessing various aspects of the data set size, however at the nodes from the lower-level of the trees meta features describing the label space appear. 
    }
    \item {\emph{Does tuning of MLC methods improves their predictive performance?}\\
We analyze the execution statistics of a large experimental study. Namely, we count the number of parameter values that were evaluated and used to select the optimal predictive model. We found that AA methods tend to explore the parameter space to a larger extent as compared to PT methods. Next, the PT methods that have SVMs as base predictive models are sensitive to the parameter values.
Finally, the optimization of the hyperparameters can improve the predictive performance, however, the extent of the improvements does not always justify the resource utilization.
    }
\end{enumerate}



In future work, we plan to extend our work in several directions. First, the meta analysis could be used to perform stratified resampling of data sets by preserving the distribution of the meta features throughout the folds. 
Next, we could design a novel method for setting the threshold when making predictions based on the meta features of a data set based on other meta features describing the properties of the label space, different from the label cardinality. Finally, we could use the meta analysis in a generative setting to generate synthetic MLC data sets for benchmarking. This is especially important for tasks like MLC the available data with good quality is still relatively scarce.

\section*{Acknowledgements}
The computational experiments presented here were executed on a computing infrastructure from the Slovenian Grid (SLING) initiative, and we thank the administrators Barbara Kra\v{s}ovec and Janez Srakar for their assistance.

\section*{Funding}
JB and DK acknowledge the financial support of the Slovenian Research Agency (research project No.~J2-9230). LT acknowledges the financial support of the Slovenian Research Agency (research program No.~P5-0093 and research project No.~V5-1930). SD acknowledges the financial support of the Slovenian Research Agency (research program No.~P2-0103) and the European Commission through the project TAILOR - Foundations of Trustworthy AI - Integrating Reasoning, Learning and Optimization (grant No. 952215). 

\bibliographystyle{acm}
\bibliography{referenceValid}

\appendix

\section{Datasets and Methods}\label{sec:datasets}

\tablename~\ref{datasetProperties} summarizes meta information for the datasets used in the study. The detailed description of both the datasets and methods can be found at \href{http://mlc.ijs.si/meta/files/IJIS_supplementary_methods_datasets.pdf}{\url{http://mlc.ijs.si/meta/files/IJIS_supplementary_methods_datasets.pdf}}.

\begin{landscape}
\begin{longtable}{l|l|r|r|r|r|r|r|r|r}
\caption{Properties of the datasets used in the study. The column "Dataset name" contains the names of the datasets. The column "Domain" contains information about the domain to which the dataset belongs to. The columns "Training" and "Test" denote the number of training and test instances available for the datasets. The column "Features" has information about the number of features the datasets consists of. The column "Labels" presents the number of labels. The columns "LCardTR" and "LCardTs" contains information about the label cardinality of the training and test set. The columns "LDeTr" and "LDeTS" contain information about the label density the training and test set have.}
\label{datasetProperties}\\

\multicolumn{1}{c|}{\textbf{Dataset name}} & \multicolumn{1}{c|}{\textbf{Domain}} & \multicolumn{1}{c|}{\textbf{Training}} & \multicolumn{1}{|c|}{\textbf{Test}} & \multicolumn{1}{c|}{\textbf{Features}} & \multicolumn{1}{c|}{\textbf{Labels}} & \multicolumn{1}{|c|}{\textbf{LCardTr}} & \multicolumn{1}{c|}{\textbf{LDeTr}} & \multicolumn{1}{c|}{\textbf{LCardTs}} & \multicolumn{1}{c}{\textbf{LDeTs}} \\ \hline \endfirsthead
\multicolumn{10}{c}%
{{\bfseries \tablename\ \thetable{} -- continued from previous page}} \\
\hline \multicolumn{1}{|c|}{\textbf{Dataset name}} & \multicolumn{1}{c|}{\textbf{Domain}} & \multicolumn{1}{c|}{\textbf{Training}} & \multicolumn{1}{|c|}{\textbf{Test}} & \multicolumn{1}{c|}{\textbf{Features}} & \multicolumn{1}{c|}{\textbf{Labels}} & \multicolumn{1}{|c|}{\textbf{LCardTr}} & \multicolumn{1}{c|}{\textbf{LDeTr}} & \multicolumn{1}{c|}{\textbf{LCardTs}} & \multicolumn{1}{c|}{\textbf{LDeTs}} \\ \hline 
\endhead
\hline \multicolumn{10}{|r|}{{Continued on next page}} \\ \hline
\endfoot
\hline
\endlastfoot
			ABPM             & Medical        &        189   &         81  &      33 &          6  &  3.9683  &  0.6614    & 3.9877  & 0.6646    \\
			Foodtruck        & Text           &        254   &        146  &      28 &         12  &  2.3268  &  0.1939    & 2.2671  & 0.1889    \\
			Flags            & Multimedia     &        174   &        110  &      19 &          7  &  3.3793  &  0.4828    & 3.4273  & 0.4896    \\
			CHD 49           & Medicine       &        372   &        183  &      49 &          6  &  2.5887  &  0.4315    & 2.5628  & 0.4271    \\

			Water quality    & Chemistry      &        711   &        349  &      16 &         14  &  5.0689  &  0.3621    & 5.0802  & 0.3629    \\
			Emotions         & Multimedia     &        413   &        270  &      72 &          6  &  1.9080  &  0.3180    & 1.8296  & 0.3049    \\   
			Virus PseAAC     & Bioinformatics &        139   &         68  &     440 &          6  &  1.2734  &  0.2122    & 1.1029  & 0.1838    \\
			VirusGo          & Bioinformatics &        139   &         68  &     749 &          6  &  1.2734  &  0.2122    & 1.1029  & 0.1838    \\
			Gpositive PseAAC & Bioinformatics &        448   &        171  &     440 &          4  &  1.0057  &  0.2514    & 1.0117  & 0.2529    \\
			Gpostive GO      & Bioinformatics &        348   &        171  &     912 &          4  &  1.0057  &  0.2514    & 1.0117  & 0.2529    \\
			Proteins Virus   & Biology        &        206   &         62  &    1538 &          6  &  1.2292  &  0.2049    & 1.1774  & 0.1962    \\ 
			Yeast            & Biology        &       1450   &        967  &     103 &         14  &  4.2566  &  0.3040    & 4.2079  & 0.3006    \\
			Birds            & Multimedia     &        445   &        290  &     260 &         19  &  1.0404  &  0.0548    & 0.9655  & 0.0508    \\
			Scene            & Multimedia     &       1447   &        960  &     294 &          6  &  1.0788  &  0.1798    & 1.0667  & 0.1778    \\
			Gnegative PseAAC & Bioinformatics &        933   &        459  &     440 &          8  &  1.0407  &  0.1301    & 1.0566  & 0.1320    \\   
			Plant PseACC     & Bioinformatics &        656   &        322  &     440 &         12  &  1.0854  &  0.0904    & 1.0652  & 0.0888    \\  
			Cal500           & Multimedia     &        362   &        230  &      68 &        174  & 25.9064  &  0.1489    &26.2223  & 0.1507    \\
			Proteins Plants  & Biology        &        673   &        288  &    1538 &         12  &  1.0847  &  0.0904    & 1.0625  & 0.0885    \\  
			Gnegative GO     & Bioinformatics &        933   &        459  &    1717 &          8  &  1.0407  &  0.1301    & 1.0566  & 0.1321    \\  
			Human PseAAC     & Bioinformatics &       2082   &       1024  &     440 &         14  &  1.1969  &  0.0855    & 1.1611  & 0.0829    \\ 
			Genbase          & Biology        &        460   &        292  &    1185 &         27  &  1.2370  &  0.0458    & 1.2568  & 0.0466    \\
			Yelp             & Text           &       7241   &       3565  &     671 &          5  &  1.6470  &  0.3294    & 1.6205  & 0.3241    \\  
			Plant GO         & Bioinformatics &        656   &        322  &    3091 &         12  &  1.0854  &  0.0904    & 1.0652  & 0.0888    \\ 
			Proteins Humans  & Biology        &       1663   &        713  &    1538 &         14  &  1.2790  &  0.0914    & 1.2637  & 0.0903    \\
			Medical          & Text           &        648   &        420  &    1499 &         45  &  1.2253  &  0.0272    & 1.2714  & 0.0283    \\
			Slashdot         & Text           &       2272   &       1513  &    1079 &         22  &  1.1787  &  0.0536    & 1.1844  & 0.0538    \\
			Enron            & Text           &       1082   &        710  &    1001 &         53  &  3.3983  &  0.0641    & 3.3366  & 0.0630    \\
			Langlog          & Text           &        940   &        610  &    1004 &         75  &  1.1489  &  0.0153    & 1.2295  & 0.0164    \\
			Arabic 200       & Text           &      15000   &       8754  &     200 &         40  &  1.1073  &  0.0277    & 1.2536  & 0.0313    \\  
			Stackexch. chess & Text        &       1005   &        670  &     585 &        227  &  2.4229  &  0.0107    & 2.3940  & 0.0106    \\
			Reutersk 500     & Text           &       3607   &       2403  &     500 &        103  &  1.4649  &  0.0142    & 1.4580  & 0.0142    \\ 
			Tmc2007 500      & Text           &      17190   &      11462  &     500 &         22  &  2.2253  &  0.1012    & 2.2107  & 0.1005    \\
			Ohsumed          & Text           &       8392   &       5592  &    1002 &         23  &  1.6621  &  0.0723    & 1.6640  & 0.0724    \\
			Ng20             & Text           &      11640   &       7750  &    1006 &         20  &  1.0288  &  0.0514    & 1.0291  & 0.0515    \\  
			Bibtex           & Text           &       4495   &       2990  &    1836 &        159  &  2.4220  &  0.0152    & 2.3625  & 0.0149    \\
			Human GO         & Bioinformatics &       2082   &       1024  &    9844 &         14  &  1.1969  &  0.0855    & 1.1611  & 0.0829    \\ 
			Stackexch. philsph & Text   &       2385   &       1593  &     842 &        233  &  2.2839  &  0.0098    & 2.2561  & 0.0097    \\
		    Stackexch. cs & Text           &       5571   &       3717  &     635 &        274  &  2.5536  &  0.0093    & 2.5582  & 0.0093    \\
			Corel5k          & Multimedia     &       3060   &       2030  &     499 &        374  &  3.5242  &  0.0094    & 3.5153  & 0.0094    \\
			Delicious        & Text           &       9725   &       6470  &     500 &        983  & 19.0851  &  0.0194    &18.9360  & 0.0193    \\ 
\end{longtable}
\end{landscape}

\end{document}